\documentclass[10pt,a4paper,twocolumn]{article}

\usepackage[a4paper,top=0.75in,bottom=0.75in,
            left=0.7in,right=0.7in,
            columnsep=0.25in]{geometry}

\usepackage{amsmath,amssymb,amsfonts}
\usepackage{mathtools,nccmath}

\usepackage{algorithm}
\usepackage{algorithmic}

\usepackage{graphicx}
\usepackage{subcaption}
\usepackage{adjustbox}
\usepackage{tikz}
\usepackage{pgf}
\graphicspath{{}}

\usepackage{multirow}
\usepackage{threeparttable}
\usepackage{makecell}
\usepackage[table]{xcolor}
\usepackage{colortbl}

\usepackage{cite}
\usepackage{textcomp}
\usepackage{balance}
\usepackage{stfloats}
\usepackage{xfp}
\usepackage{fancyhdr}
\usepackage{float}

\usepackage[hidelinks]{hyperref}

\usepackage{setspace}

\def\BibTeX{{\rm B\kern-.05em{\sc i\kern-.025em b}\kern-.08em
T\kern-.1667em\lower.7ex\hbox{E}\kern-.125emX}}

\def\CM#1{ {\color{black} {#1}}}  
\def\cdm#1{ {\color{black} {#1}}}

\newenvironment{cm}
{\par\begingroup\color{black}}
{\par\endgroup}

\newcommand{\encoderparamimg}{$\phi_{\mathcal{E}}^i$}
\newcommand{\encoderparamtext}{$\phi_{\mathcal{E}}^t$}
\newcommand{\decoderparam}{$\phi_{\mathcal{D}}$}
\newcommand{\encoderimg}{$\mathcal{F}_{\phi_{\mathcal{E}}^i}$}
\newcommand{\encodertext}{$\mathcal{F}_{\phi_{\mathcal{E}}^t}$}
\newcommand{\encoder}{$\mathcal{F}_\mathcal{E}$}
\newcommand{\decoder}{$\mathcal{F}_{\phi_{\mathcal{D}}}$}

\newcommand{\colorrampred}[3]{%
  \begingroup%
  \pgfmathparse{int(100*(#1 - #2)/(#3 - #2))}%
  \edef\x{\endgroup\noexpand\cellcolor{red!50!white!\pgfmathresult}#1}%
  \x%
}

\newcommand{\colorrampgreen}[3]{%
  \begingroup%
  \pgfmathparse{int(100*(#1 - #2)/(#3 - #2))}%
  \edef\x{\endgroup\noexpand\cellcolor{green!50!white!\pgfmathresult!}#1}%
  \x%
}

\begin{document}


\onecolumn

\begin{titlepage}
\thispagestyle{empty}

\begin{center}

\vspace*{1.5cm}

{\LARGE\bfseries
TransXplainRRG: A Radiologist-Evaluated LVLM
for Chest Radiograph Report Generation
\par}

\vspace{1.2cm}

{\large
Chayan Mondal$^{1,2}$,
Duc-Son Pham$^{1}$,
Tele Tan$^{1}$,
Ashu Gupta$^{2}$,
Vincent T.Y. Tiong$^{2}$,
Shreya Ghosh$^{1}$,
and Tom Gedeon$^{1}$
\par}

\vspace{0.8cm}

{\normalsize
$^{1}$Curtin University, Bentley, Western Australia\\
$^{2}$Fiona Stanley Hospital, Murdoch, Western Australia
\par}

\vspace{0.8cm}

{\normalsize
Corresponding author: Chayan Mondal\\
Email: chayan.deb@ieee.org
\par}

\vspace{50pt}

\begin{center}
\begin{minipage}{0.9\textwidth}
\centering
\fontsize{10}{13}\selectfont
This is an unofficial, extended author-prepared version of the article
accepted in the IEEE Journal of Biomedical and Health Informatics.
The official version of record is available on IEEE Xplore.
This version includes supplementary explanations and an appendix
omitted from the published article to meet page-length constraints.
Please cite the official published version.
\end{minipage}
\end{center}

\vfill

\begin{minipage}{0.9\textwidth}
\large
\textbf{Copyright Notice}

© 2026 IEEE. Personal use of this material is permitted.
Permission from IEEE must be obtained for all other uses,
in any current or future media, including reprinting/
republishing this material for advertising or promotional
purposes, creating new collective works, for resale or
redistribution to servers or lists, or reuse of any
copyrighted component of this work in other works.
\end{minipage}

\end{center}
\end{titlepage}


\twocolumn

\setcounter{page}{1}

\twocolumn[
\begin{center}
{\Large\bfseries
TransXplainRRG: A Radiologist-Evaluated LVLM
for Chest Radiograph Report Generation
\par}

\vspace{0.4cm}

{\normalsize
Chayan Mondal, Duc-Son Pham, Tele Tan, Ashu Gupta,
Vincent T.Y. Tiong, Shreya Ghosh, and Tom Gedeon
\par}

\vspace{0.4cm}
\end{center}
]

\begin{abstract}
Recent advancements in Large Vision-Language Models (LVLM) enhance X-ray report generation, but often neglect clinical context, overlook workflow integration, and lack explainability, which hinders clinical adoption due to interpretability and reliability concerns. This paper addresses the dual challenges of performance and expert explainability in automated X-ray report generation for clinical AI applications. Our proposed approach, TransXplainRRG, utilises an off-the-shelf Swin Transformer model along with a transformer-based text encoder that incorporates patient medical history to generate a radiology report. Further, we explore an expert-guided `inside-out' approach and extract only abnormal findings for radiology report refinement. Thus, this study bridges the gap between high-performance automation and interpretability critical for clinical practice by combining state-of-the-art transformer-based vision encoders, text encoders, and LLMs. TransXplainRRG is trained and tested on the large-scale MIMIC-CXR dataset and further evaluated on the unseen IU X-Ray dataset to demonstrate its generalizability and robustness. \cdm{We employed the GREEN metric as a complementary measure of clinical consistency, achieving scores of 0.327 on MIMIC-CXR and 0.605 on IU X-Ray, both higher than those achieved by MAIRA-2 under the same evaluation setting.} It also performed competitively across standard evaluation metrics, frequently matching or surpassing state-of-the-art models. 
Additionally, we introduce qualitative evaluation metrics developed from radiologists' viewpoints \cdm{and conduct a preliminary radiologist evaluation on 100 generated reports to assess their clinical relevance.}
Quantitative and qualitative analyses further \cdm{demonstrate the potential of} our TransXplainRRG framework \cdm{for reliable and explainable radiology report generation.}
\end{abstract}


\section{Introduction}
\label{sec:introduction}
Chest radiographs are vital for clinical decision-making, providing essential diagnostic and prognostic insights. Interpreting these images requires both clinical expertise and considerable time, contributing to the growing workload of radiologists, particularly amid global shortages. Recent advances in machine learning have paved the way for automated Radiology Report Generation (RRG), where Vision-Language Models (VLMs) learn to translate image features into clinically meaningful textual descriptions. Automated RRG systems have the potential to reduce radiologists’ repetitive workload, standardise report quality, and improve communication within clinical teams~\cite{ rajpurkar2023current}. Moreover, they can assist in triaging cases, enhancing reporting efficiency, and minimizing delays in patient care.

In this work, we propose a clinically informed RRG framework that integrates powerful visual encoders for disease classification and large language models (LLMs) for generating accurate and context-aware diagnostic reports. Our approach advances prior work by systematically identifying optimal visual backbones and designing a generation pipeline that is explainable, robust, and aligned with clinical needs. This unified model not only supports high-quality report generation but also offers a step toward trustworthy AI-assisted decision support in radiology workflows. 

Significant research has focused on designing automatic report-generation systems. However, these systems often overlook the procedures typically followed by radiologists during the report generation process. 
Radiologists do not examine chest X-rays in isolation but interpret them in the context of doctors' requests, following a systematic approach \cite{de2012chest} such as the inside-out method (see Fig.~\ref{fig:Systematic_approach}). This method begins with assessing the heart, mediastinum, and hili, followed by the lungs, lung borders, chest wall, and abdomen, ensuring all interfaces with the diaphragm, heart, and mediastinal structures are examined. Familiarity with normal anatomy and variations is crucial. Mediastinal lines and silhouette signs help detect subtle abnormalities, after which a pattern-based approach aids in diagnosing and considering alternatives~\cite{CXRInterpretation2022}. 
Whereas previous work often emphasises generating detailed reports, creating concise summaries, known as impressions, can significantly enhance radiologists' efficiency and help address workload challenges by enabling quicker, informed decision-making.
AI-based systems that identify and highlight abnormalities in chest X-ray reports can greatly benefit radiologists by improving diagnostic accuracy and efficiency. These systems streamline workflows by emphasising critical findings, enabling radiologists to identify issues without unnecessary details quickly. This targeted assistance accelerates report review and correction, ensuring timely and accurate diagnostics. As a result, AI-driven tools enhance radiological efficiency and effectiveness, ultimately improving patient care. The widespread ability of different LLMs motivated us to utilise them in refining generated radiology reports, focusing on a systemic inside-out approach and highlighting only the abnormal findings from the generated report. \CM{Rather than proposing a new backbone architecture, this work introduces a clinically grounded and disease-aware vision language report generation framework that integrates optimised visual encoding, topic-guided text encoding, state-aware multimodal fusion, and risk-based clinical evaluation.} Our main contributions to this paper are as follows:
\begin{itemize}
       
    \item We propose TransXplainRRG, an LVLM model that encodes CXR images alongside patients' medical history by integrating LLMs at the output stage to enhance radiology report generation.
   
    \item We further incorporate prompt engineering techniques to structure radiology reports for clinical purposes, maintaining an `inside-out' approach and filtering for abnormal findings only to assist radiologists' clinical workflow in chest X-ray report generation.
    
    \item We propose qualitative evaluation metrics designed from radiologists' perspectives to assess the clinical relevance and utility of the generated reports in real-world settings.
    
    \item We demonstrate comprehensive quantitative and qualitative analysis on MIMIC-CXR and IU X-ray datasets to show the robustness of our proposed method. 
\end{itemize}

\section{Related Work}
\noindent \textbf{Visual Encoder for Radiology Reports.} A deep learning-based automatic report generation system for medical imaging integrates vision and text processing modules. Popular vision encoder architectures include CNN~\cite{mondal2023efficientransnet}, and Transformer~\cite{park2022self} models that classify an input X-ray image to corresponding disease classes. TransXplainNet, a fully transformer-based automatic report generation system~\cite{mondal2023transformers}, employs a DeiT-CXR model as a vision encoder to classify 14 common findings in chest X-ray images. The authors pre-trained the DeiT model on the domain-specific CheXpert dataset to enhance feature extraction of CXR images, which improved their report generation approach through better image understanding. 

Another variation of ViT is the Swin Transformer \cite{liu2021swin}, which introduces a hierarchical vision transformer utilizing shifted windows. This model addresses the challenges in adapting Transformers from language to vision by employing a shifted windowing scheme. This approach limits self-attention computations to non-overlapping local windows, enhancing efficiency while maintaining cross-window connections. The hierarchical design allows the Swin Transformer (Swin-T) to effectively handle visual entities of various scales and achieve linear computational complexity with respect to image size. Its strong performance across tasks such as image classification, object detection, and segmentation \cite{liu2024swin}, as well as virtual fluorescence staining in medical imaging, underscores its versatility as a general-purpose backbone for computer vision.

\noindent \textbf{Medical Report Generation.} 
\CM{Early work on automatic medical report generation primarily relied on CNN-RNN-based encoder-decoder frameworks to translate visual information into textual descriptions. Shin et al. \cite{shin2016learning} explored disease identification and image description using MeSH annotations derived from radiology reports, demonstrating that deep CNN encoders combined with recurrent decoders could generate meaningful medical text from images. Building on this paradigm, TieNet \cite{wang2018tienet} introduced an attention-guided CNN-RNN architecture for chest X-ray report generation, enabling the model to attend to salient image regions while producing clinically relevant descriptions. Although these early approaches established the feasibility of end-to-end report generation, they were limited in capturing long-range dependencies and complex clinical context.}

\CM{Subsequent studies shifted toward transformer-based architectures to improve report coherence and global reasoning. Several works combined CNN-based visual feature extractors with transformer encoder-decoder models to better capture contextual dependencies in both image and text modalities. For example, DenseNet-based transfer learning has been used to extract global and local visual features that are later processed as token sequences by transformer encoders for report generation~\cite{amjoud2021automatic}. To improve semantic consistency, Nguyen et al. \cite{nguyen2021automated} proposed an interpreter module that integrates clinical history and imaging information to construct enriched disease embeddings prior to report generation. In parallel, pretrained language models such as GPT-2 have been explored as decoders, guided by disease tags predicted from CNN encoders~\cite{alfarghaly2021automated}. While these approaches improved fluency and semantic consistency, they largely focused on architectural refinement and did not explicitly model radiologist reporting workflows or abnormality prioritisation.}

More recent research has moved beyond purely architectural improvements toward structured supervision and explicit cross-modality alignment. CAMANet~\cite{wang2024camanet} promotes region-level visual-textual alignment by introducing class activation map supervision to guide report generation. A more recent framework~\cite{zhang2025automatic} incorporates Self-Mamba and Cross-Mamba modules to enhance global feature modelling and cross-modal interaction, together with a Sparse-Loss function to address sample imbalance. These approaches highlight the importance of structured constraints and cross-modal mechanisms for improving abnormality-aware report generation.

\CM{Large-scale multimodal models have further advanced radiology report generation by jointly addressing visual understanding, report generation, and clinical interpretability. MAIRA-2~\cite{bannur2024maira} is a large multimodal model for automated chest X-ray report generation that supports visual grounding, linking reported findings to corresponding image regions. It also introduces RadFact, an LLM-based framework for fine-grained evaluation of sentence-level correctness and completeness. In parallel, Nicolson et al.~\cite{nicolson2024longitudinal} explored longitudinal report generation by incorporating prior and non-prior studies, using section embeddings and a CXR-BERT-based reinforcement learning reward to improve clinical semantic consistency.}


\cdm{Beyond report generation, recent work has explored human-centred interpretability by integrating human visual attention with VLM-generated semantic explanations. Chen et al.~\cite{chen2025gaze} proposed a teacher-student framework for weakly supervised medical image segmentation that combines clinician gaze indicating diagnostically relevant regions, with VLM-generated descriptions that provide semantic and morphological context. This work highlights the value of aligning model explanations with clinically relevant regions and human reasoning. In contrast, TransXplainRRG focuses on chest X-ray report generation and combines Grad-CAM-based visual attribution with structured LLM-based report refinement to provide complementary visual and textual explanations aligned with radiology reporting.}


\noindent \textbf{LVLM, LLM and Medical Chatbot.} 
Recent studies have demonstrated the potential of large vision-language models (LVLMs) for tasks such as image captioning and visual question answering~\cite{bazi2023vision}. LLMs, including GPT-4 and Llama~\cite{touvron2023llama}, have further advanced multimodal understanding and text generation in medical applications. These developments motivate the use of LLMs for structured and clinically oriented radiology report refinement. Medical chatbots have also emerged as promising healthcare tools, highlighting the importance of reliability and safety~\cite{li2023chatdoctor}.


\cdm{Despite progress, including recent efforts toward human-centred visual and semantic interpretability, many existing approaches still lack explicit alignment with radiologists' systematic reporting practices, risk prioritisation, and abnormality-focused summarisation, which are critical for efficient decision making in high-pressure clinical environments. To address these limitations, TransXplainRRG incorporates patient medical history to provide clinically relevant context, adopts an expert-guided `inside-out' approach to prioritise abnormal findings during report refinement, and combines visual attribution with structured LLM-based refinement to enhance explainability. They aim to provide a clinically grounded and workflow-aligned framework for radiology report generation, while also considering practical deployment requirements.}

\section{Method}

\begin{figure*}[!ht]
  \centering
 \includegraphics[width=0.95\linewidth]{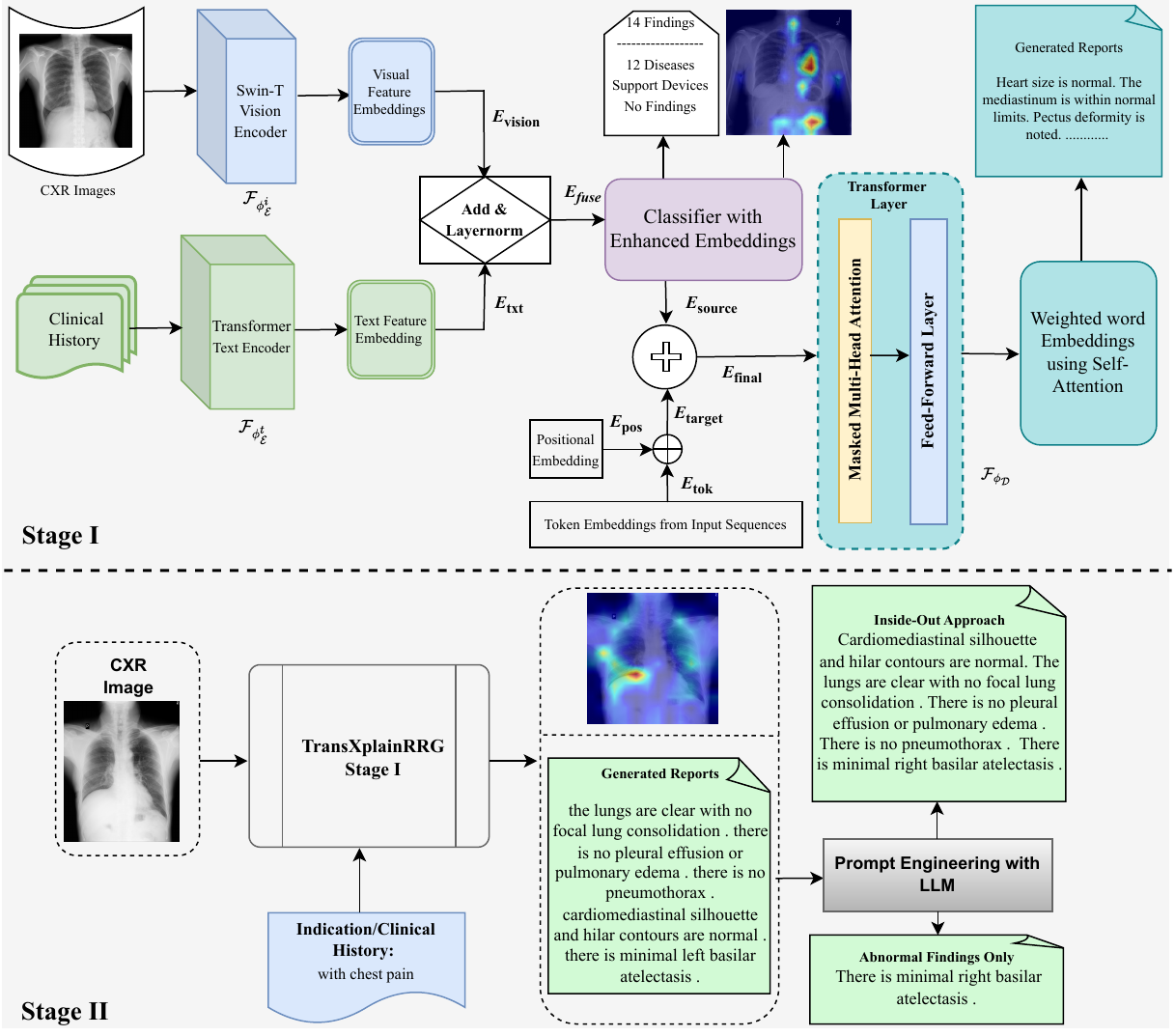}
  \caption{Proposed two-stage TransXplainRRG framework for generating and refining chest X-ray reports using clinical history and large language model integration.}
  \vspace{-20pt}
  \label{fig:Block_diagram}
\end{figure*}

\textit{TransXplainRRG} (see Fig.~\ref{fig:Block_diagram}) is a \textit{two-stage} vision-language model specifically designed to analyse chest radiographs to assist medical professionals with insightful and efficient report generation. \CM{Unlike approaches that rely purely on off-the-shelf multimodal architectures, the proposed framework introduces a clinically grounded vision-language design that integrates disease-topic-guided text encoding, disease-state-aware multimodal fusion, and structured report refinement. These components collectively enable clinically meaningful report generation while maintaining computational efficiency and deployment feasibility in real-world clinical environments.} In the \textit{first stage}, a pre-trained medical vision encoder (that is, a standard Swin-Transformer backbone) and a transformer encoder simultaneously align visual and textual representations through the add \& normalization layer. The proposed multi-objective learning formulation maps the input CXR image to the optimal output representation manifold, which can further be utilised for clinically \cdm{relevant} and efficient report generation. In the \textit{second stage}, prompt engineering with a large language model is applied to refine the reports generated from the first stage in a consistent structural format. For reproducibility, the implementation of our model is made available at \href{https://github.com/Chayaneee/TransXplainRRG}{https://github.com/Chayaneee/TransXplainRRG}.

\vspace{-10pt}
\subsection{TransXplainRRG Stage I}

Our TransXplainRRG model $\mathcal{F}$ consists of an image encoder (\encoderimg), text encoder (\encodertext) and decoder (\decoder) with embedding parameters \encoderparamimg, \encoderparamtext\ and \decoderparam\, respectively. \CM{The proposed framework adopts a task-driven vision-language encoder-decoder design, 
where multimodal alignment is achieved implicitly through supervised report generation 
rather than explicit contrastive embedding learning.} The encoder model $\mathcal{F}_\mathcal{E}$ consists of joint learning of two sub-networks: $(\mathcal{F}_{\phi_{\mathcal{E}}^i} \circ \mathcal{F}_{\theta})$ and $(\mathcal{F}_{\phi_{\mathcal{E}}^t} \circ \mathcal{F}_{\theta})$.

\noindent \textbf{Image Encoder} (\encoderimg). Our published study \cite{mondal2024pursuing} conducted an ablation analysis to evaluate the effectiveness of the Swin Transformer (Swin-T) for classifying the life-threatening condition Pneumothorax. Based on its superior performance, we adopted Swin-T as the vision encoder in our TransXplainRRG pipeline. An input CXR image $x_i\in \mathbb{R}^{H \times W \times 3}$ with frontal and lateral view is divided into small, fixed-size, non-overlapping patches ($\mathbb{R}^{4 \times 4}$), resulting in each patch having a feature dimension of ($x_i^p\in \mathbb{R}^{4 \times 4 \times 3}$) to give input to a Swin-T encoder (\encoderparamimg).
The patches are mapped to a high-dimensional space and projected to a specified dimension $C$ through a linear embedding layer, creating a sequence of embedded patch tokens. These tokens are then passed through the four stages of the Swin-T architecture. At each stage transition, a patch merging layer combines adjacent patches, reducing the token count, down-sampling the spatial resolution, and simultaneously increasing the feature dimension. Finally, the image is mapped to the feature embedding space by \encoderimg: $x_i \to E_{vision}$, where $E_{vision} \in \mathbb{R}^{n \times e}$. Here, $n=114$ denotes the number of disease-related topics, and $e=256$ is the embedding dimension. A learnable linear projection maps the extracted visual features to these 114 topic embeddings, aligning them with the text representations for subsequent multimodal fusion.

\noindent \textbf{Text Encoder} (\encodertext).
Clinical history documents and chest X-ray images are used to integrate relevant medical information into the radiology workflow. 
The clinical history text ${x_t}\in \mathbb{R}^{l}$ is mapped to the feature embedding space by \encodertext$:x_t \rightarrow E_{txt}$, where $l$ denotes the number of tokens and $E_{txt}$ represents the compact disease-aware text embedding.

The transformer-based text encoder first converts the clinical history into token embeddings enhanced with positional encodings to preserve sequential order. These embeddings are passed through multiple transformer layers consisting of multi-head self-attention and feedforward networks to capture contextual relationships between tokens. The encoder's output is a sequence of contextual hidden states $H \in \mathbb{R}^{l \times e}$.

\begin{figure}[t]
  \centering
  \includegraphics[width=\linewidth]{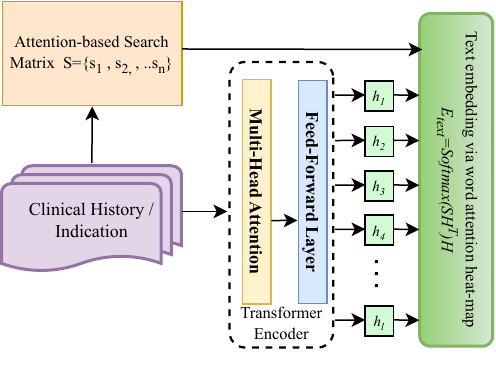}
  \caption{\CM{Architecture of the text encoder and attention-based search matrix used to extract disease-relevant contextual information from clinical history. The transformer encoder produces contextual hidden states, while the search matrix guides the model to focus on disease-specific topics and generate compact text embeddings.}}
  \vspace{-5pt}
  \label{fig:text_encoder_search}
\end{figure}


\cdm{Figure~\ref{fig:text_encoder_search} illustrates the text encoder and attention-based search matrix used to extract disease-relevant contextual information from clinical history. To improve report-generation coverage, the search matrix is constructed from 14 disease labels and 100 frequent noun phrases extracted from the training reports. These 114 disease-related topics are represented as learnable queries:
$S=\{s_1,s_2,\ldots,s_n\}$, where $S\in\mathbb{R}^{n\times e}$ and $n=114$. The queries attend to the contextual hidden states to extract topic-specific information from the clinical history.}

The disease-aware text embedding is computed using an attention mechanism:

\begin{equation}
E_{txt} = \texttt{softmax}(SH^{T})H
\label{eq:text_embedding}
\end{equation}

\noindent where the softmax operation produces a word-attention map that highlights the tokens most relevant to each disease topic. The resulting weighted sum generates the final disease-aware textual representation $E_{txt} \in \mathbb{R}^{n \times e}$. This embedding provides a concise, disease-specific feature representation of the clinical history, which can be integrated with chest X-ray image features to generate a comprehensive radiology report.

\noindent\textbf{Classifier with Enhanced Embeddings}. (\encoder).
The embedded features $E_{vision}$ and $E_{txt}$, representing visual and textual information, are fused using a residual fusion mechanism:

\begin{equation}
E_{fuse} = \text{LayerNorm}(E_{vision} + E_{txt})
\label{eq:fusion}
\end{equation}

\CM{This lightweight fusion strategy preserves modality-specific information while enabling joint multimodal representation learning. These combined features, $E_{fuse}$, are mapped to the output space $y\in \mathbb{R}^{14}$ via the function $\mathcal{F}_{\theta}:E_{fuse} \to y$. To incorporate disease-state awareness, each disease category is associated with learnable state embeddings representing positive, negative, and uncertain conditions:}

\CM{\begin{equation}
E_{state} = \{e_{pos}, e_{neg}, e_{unc}\}, \quad e_i \in \mathbb{R}^{e}
\label{eq:state_embedding}
\end{equation}}

\CM{\noindent where each state embedding $e_i$ captures a disease-specific semantic representation for its corresponding clinical state. These state embeddings interact with the fused multimodal representation through an attention mechanism to produce a state-aware representation:}

\CM{\begin{equation}
E_{states} = \text{Attention}(Q=E_{fuse},K=E_{state},V=E_{state})
\label{eq:state_attention}
\end{equation}}

\CM{This mechanism allows the model to dynamically attend to disease-state information when interpreting the fused visual and textual features, enabling the model to represent both the presence and uncertainty of clinical findings.}

The disease state-aware embedding $E_{states}$ classifies diseases into three states, guided by ground-truth labels during training and predicted via attention-weighted features during inference. The classifier generates the final output, including attention weights and $E_{states}$, and constructs an enhanced embedding $E_{source}$ by combining disease class information, $E_{states}$, and $E_{fuse}$. This enriched embedding is used by the decoder (\decoder) to produce clear, accurate disease descriptions, supporting the overall goal of the classifier to provide meaningful and actionable insights.

\noindent \textbf{Decoder (\decoder).} In the report generation pipeline, the decoding block builds on the Transformer decoder framework, with each of its \emph{N} layers consisting of a masked multi-head self-attention mechanism followed by a feedforward network. This process begins by embedding each target token with a positional embedding, producing 
$E_{target}$ = $E_{token}$ + $E_{pos}$. This target embedding is concatenated with source embeddings 
$E_{source}$ from the encoder, resulting in the combined input 
$E_{final}$ = [$E_{source}$, $E_{target}$]
, which initialises the decoding stage. Within each decoder layer, the masked multi-head self-attention allows each token position, represented in the hidden states 
{\emph{H}} = $[h_{1},h_{2}, h_{3},...,h_{m}]$, to attend exclusively to itself and previous tokens, preserving the autoregressive nature of report generation. This attention, directed by the enhanced disease embedding $E_{source}$ derived from the encoder stage, enhances the clinical relevance of generated tokens.

Each layer processes $E_{final}$ through an attention mechanism regulated by an attention mask $M_{att}$ that restricts future tokens from influencing the current token's prediction. The resulting attention output refines the contextual representation of each token, which is then passed through a feedforward network to deepen the embedding. After $N$ layers, this sequence yields refined embeddings $H_{final}$, employed to predict the next token in the generated report. These predictions use a vocabulary embedding 
$V\in \mathbb{R}^{v \times e}$, where 
\emph{v} denotes vocabulary size, enabling weighted word embeddings $W\in \mathbb{R}^{m\times e}$ through $W  = \texttt{softmax} (HV^T)V $. 



\cdm{The model is trained using a combined cross-entropy loss:
\begin{equation}
\mathcal{L}_{\mathrm{total}}
=
\mathcal{L}_{\mathrm{CE}}^{\mathrm{shift}}
+
\mathcal{L}_{\mathrm{CE}}^{\mathrm{aux}},
\label{eq:total_loss}
\end{equation}
where $\mathcal{L}_{\mathrm{CE}}^{\mathrm{shift}}$ is the one-token-shifted report-generation loss and $\mathcal{L}_{\mathrm{CE}}^{\mathrm{aux}}$ is the auxiliary classification loss. Both terms are equally weighted. The report-generation loss uses $\texttt{ignore\_index}=3$, corresponding to the padding token (\texttt{<pad>}), so padded positions are excluded.} This iterative decoding method enables structured and accurate report generation by dynamically integrating source and target sequence information.

\CM{It is important to clarify that Stage I focuses on hierarchical multimodal representation learning and contextual report generation, while anatomical inside-out structuring is introduced later during Stage II refinement rather than being explicitly enforced within the vision encoder.}

\subsection{TransXplainRRG Stage II}
\CM{\noindent \textbf{Report refinement via large language model (LLM).} 
The block diagram in Stage II of Fig.~\ref{fig:Block_diagram} presents a refinement module that operates on the reports generated by TransXplainRRG Stage I. After training the Stage I model, a complete radiology report is generated from the chest X-ray image and clinical history. Grad-CAM visualisations are also produced to highlight image regions contributing to the generated findings, improving interpretability for clinical users.}

\CM{It is important to note that all quantitative benchmarking results reported in this study are obtained exclusively from the outputs of TransXplainRRG Stage I. Stage II operates as a downstream refinement step applied during inference to improve report organisation, interpretability, and clinical usability rather than automatic evaluation scores.} \CM{Building on the Stage I outputs, prompt-guided large language models are used to refine the generated reports into clinically meaningful formats. Two structured report representations are explored.}

\CM{\noindent \textbf{1. Abnormal findings-only reports:} This structure extracts only clinically relevant abnormal findings from the generated report. Such summaries allow rapid identification of important observations in time-critical clinical environments. The refinement is performed using prompt-guided LLMs to filter out normal statements, retain clinically significant abnormalities, and support device descriptions.}

\begin{figure}[t]
  \centering
  \includegraphics[width=\linewidth]{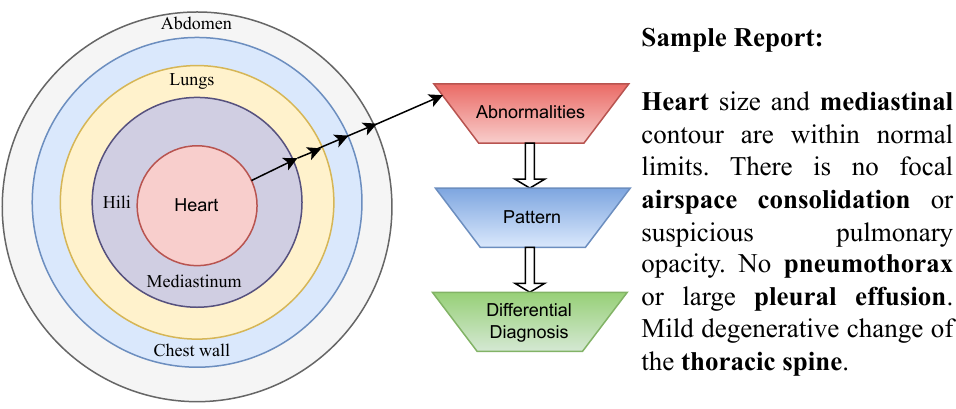}
  \caption{The systematic inside-out approach of practical chest X-ray examination.}
  \vspace{-5pt}
  \label{fig:Systematic_approach}
\end{figure}

\CM{\noindent \textbf{2. Inside-Out Approach-Based Reports:}
Inspired by practical radiological examination workflows, the refined report follows a systematic inside-to-outside organisation illustrated in Fig.~\ref{fig:Systematic_approach}. The report first describes support devices and then progresses anatomically from thoracic structures (e.g., heart and mediastinum) toward surrounding regions (e.g., lungs and abdomen). This organisation improves readability and facilitates faster clinical interpretation.}

\CM{To support practical deployment, the framework explores both commercial LLMs and small-scale open-source models such as Llama3-8B. Instead of model fine-tuning, carefully designed prompts guide the refinement process, enabling efficient report restructuring with minimal computational overhead.}

\section{Experiments}

\begin{figure}[t]
  \centering
  \includegraphics[width=\linewidth]{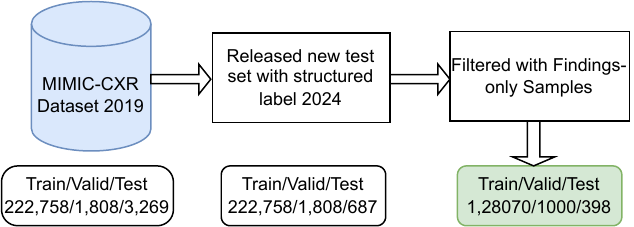}
  \caption{Preparation pipeline of the MIMIC-CXR dataset. Unlike many prior works that used earlier versions, we evaluate on the updated 2024 test set with radiologist-verified labels, providing ground truth for more robust real-world assessment. The train/validation/test splits at each stage are shown below the corresponding blocks.}
  \label{fig:mimic-split}
\end{figure}

\noindent \textbf{Datasets.} We utilised two publicly available datasets mentioned below: 

\noindent \textit{1. MIMIC-CXR.}
The MIMIC-CXR dataset \cite{johnson2019mimic} is a large-scale collection of 377,110 chest X-ray (CXR) images and 227,835 detailed medical reports from 65,379 patients, structured with sections, including clinical history, findings, and impressions. Annotated using the CheXpert labeller \cite{liu2019clinically} with 14 observation categories, it provides ground truth labels. For our study, we used the official train/validation split from MIMIC-CXR, but evaluated on the updated test set released in 2024 \cite{johnson2024mimic}, which contains human annotations from a single radiologist used for evaluating labelling performance. Studies with a missing \texttt{FINDINGS} section were excluded, and a maximum of two images per study were retained from the official training and validation splits. This gave a split of 51,657 / 422 / 213 unique patients, 128,070 / 1000 / 398 studies, and 196,415 / 1530 / 579 CXR images for Train / Validation / Test sets. The dataset split is presented in Figure \ref{fig:mimic-split}. While many prior works evaluated on the earlier test set, we used the updated 2024 version with radiologist-verified labels, as it offers \cdm{radiologist-verified} ground truth to better assess model robustness in real-world settings. 

\noindent \textit{2. IU-CXR.} The IU X-ray dataset \cite{demner2016preparing} comprises 3,955 radiology examinations, encompassing a total of 7,470 chest X-ray images that include both frontal and lateral views. Each examination corresponds to a unique patient. The associated reports typically contain sections such as findings, impressions, comparisons, and indications. Since the dataset does not provide explicit clinical history, we use the indications section to supply contextual information for our model. We evaluate our model using the official train/validation/test split, where the test set contains 668 cases. Additionally, to assess the generalizability and robustness of our model, we use the entire dataset (3,337 samples) as a test set, applying model weights trained on the MIMIC-CXR dataset.

\noindent \textbf{Evaluation Metrics.}
Traditional word-overlap metrics such as BLEU, METEOR, ROUGE-L, and CIDEr are widely used in natural language generation and report generation methods \cite{wang2024camanet, zhang2025automatic} but fall short in assessing the clinical accuracy of radiology reports. To address this, Clinical Efficacy (CE) \cite{liu2019clinically} was introduced, using the CheXpert labeller \cite{irvin2019chexpert} to extract 14 medical observations and compute scores like precision, recall, AUC, and F1 between generated and reference reports. More recent radiology-specific metrics include F1-ChexBERT \cite{smit2020chexbert}, which measures overlap in clinical findings using a BERT-based classifier trained on CheXpert labels, and F1-RadGraph \cite{delbrouck-etal-2022-improving}, which evaluates the overlap of structured clinical entities and relations.

To capture clinical correctness and improve interpretability, we adopt the GREEN (Generative Radiology Report Evaluation and Error Notation) metric~\cite{ostmeier2024green}. GREEN uses language models to identify and explain clinically significant errors, providing both a quantitative score and interpretable feedback. It correlates well with expert assessments and offers a clinically oriented evaluation of generated radiology reports.

To complement these automatic evaluation metrics, we additionally propose a radiologist-led qualitative framework to assess clinical safety from a real-world perspective. A board-certified radiologist reviewed 100 reports generated by our proposed model to assess clinical safety. The reports were categorised into four criteria: Immediate Risk (IR), representing reports that could result in immediate adverse outcomes, such as a missed pneumothorax; Long-term Risk (LTR), capturing reports that miss incidental findings with potential future health implications, such as undetected cardiomegaly; Combined Risk (CR), representing reports containing both immediate and long-term risks; and No Risk (NR), indicating clinically accurate reports without any identifiable risk.

\CM{These categories are not intended to represent standardised clinical diagnostic labels. Instead, they were designed as a safety-oriented evaluation framework to capture the potential clinical impact of errors in report generation. The definitions of these categories were formulated in consultation with radiologists to reflect clinically meaningful levels of patient risk associated with missed or incorrectly reported findings. This human evaluation, therefore, provides an essential complement to automatic metrics, focusing on patient safety and the clinical implications of errors, thereby reinforcing the real-world applicability of our system. To further assess the reliability of this qualitative evaluation framework, a second independent radiologist reviewed the same set of reports, and inter-rater reliability was quantified using agreement and Cohen’s $\kappa$ statistics.}

Further qualitative assessment was performed after using prompt engineering with different LLMs on the MIMIC test set based on the following structures:

\noindent \textit{1. Abnormal Findings-only Reports}.
Abnormal findings-only reports were qualitatively assessed by clinicians using two approaches. First, erroneous findings were scored from 0 to 3: Score 0 indicated all abnormal findings were correctly identified, Score 1 indicated missing findings, Score 2a indicated false findings, Score 2b indicated normal findings misclassified as abnormal, and Score 3 indicated both missing and false findings. Second, clinical safety was evaluated as either safe (Score 1), where no critical findings were missed and patient care remained unaffected, or unsafe (Score 0), where missing or false findings could impact diagnosis or treatment.

\noindent \textit{2. Inside-out approach}. 
\label{EM_In_Out}
The quality of inside-out approached reports is assessed based on the degree of error, scored from 0 to 4. A perfect report (Score 0) maintains the correct sequence and includes all findings. Scores increase with specific issues: Score 1 for altered order, Score 2 for added findings, Score 3 for missing findings, and Score 4 for omitted critical abnormal findings that could impact patient care.

\noindent \textbf{Implementation Details.}
We employed the AdamW optimiser with an initial learning rate of 0.0003.\cdm{To optimise training, we employed a MultiStepLR scheduler with a decay factor of $\gamma=0.1$, applied at epoch 5 for MIMIC-CXR and epoch 25 for IU X-ray.} The experiments were carried out with a batch size of 16 and an image resolution of $384 \times 384$. To avoid overfitting, we applied a weight decay of 0.02 and a dropout rate of 0.1 during training.\cdm{The transformer-based decoder uses a 256-dimensional embedding space, a feed-forward dimension of 256, and 12 transformer layers with a single attention head. The clinical-history search matrix is implemented as a learnable embedding matrix containing 114 disease-related topic queries. Its weights are initialised using PyTorch's default embedding initialisation.} All experiments were performed using PyTorch on a Linux server equipped with an NVIDIA GeForce RTX 4090 with 24GB of GPU memory. The computational complexity is dominated by the vision encoder, text encoder, and transformer-based decoder during training. The LLM is used only at inference time for report structuring and does not require fine-tuning, thereby avoiding the computational and memory overhead of LLM fine-tuning.\cdm{The complete Stage II prompts are provided in the Table~\ref{tab:Prompt_abnormal}, and Table~\ref{tab:Prompt_Inside_Out}  of Appendix~\ref{Prompt_abnormal}.}

\section{Results}

\subsection{Ablation Study}

\cdm{We investigated the contribution of the key components of the proposed TransXplainRRG framework via an ablation study using the IU X-ray dataset. The effects of the clinical-history branch with the proposed search matrix, the multimodal fusion strategy, and the state embedding module were evaluated. Table \ref{tab:ablation_study} shows the BLEU-4, ROUGE-L, F1-CheXBert, and F1-RadGraph scores to assess the quality and clinical consistency of the generated reports.}

\cdm{The image-only variant, in which the clinical-history branch and associated search matrix were removed, achieved BLEU-4, ROUGE-L, F1-CheXBert, and F1-RadGraph scores of 0.154, 0.335, 0.390, and 0.285, respectively. Replacing the proposed Add + LayerNorm fusion with Cross-Attention resulted in lower BLEU-4 (0.139), ROUGE-L (0.332), and F1-CheXBert (0.382), while a comparable F1-RadGraph score of 0.286 was obtained. A more substantial performance degradation was observed when the state embedding module was disabled, with the corresponding scores decreasing to 0.108, 0.303, 0.361, and 0.249, respectively. In contrast, the complete TransXplainRRG achieved the best performance across all four metrics, obtaining 0.228 BLEU-4, 0.438 ROUGE-L, 0.655 F1-CheXBert, and 0.337 F1-RadGraph. These results demonstrate that the search-matrix mechanism, the proposed Add + LayerNorm fusion, and the state embedding module collectively contribute to TransXplainRRG's performance.}

\begin{table*}[ht]
\centering
\caption{\cdm{Ablation study of the proposed TransXplainRRG framework. The image-only variant removes the clinical-history branch (including the proposed search matrix), Cross-Attention is compared with the proposed Add + LayerNorm fusion, and the contribution of the state embedding module is evaluated by disabling state embeddings. }}
\label{tab:ablation_study}
\begin{adjustbox}{width=\linewidth}
\begin{tabular}{|l|c|c|c|c|c|c|c|c|}
\hline
\textbf{Model} &
\textbf{Clinical} &
\textbf{Search Matrix} &
\textbf{Fusion} &
\textbf{State} &
\textbf{B4} &
\textbf{ROUGE-L} &
\textbf{F1-CheXBert} &
\textbf{F1-RadGraph} \\
&
\textbf{History} &
\textbf{(Text Encoder)} &
&
\textbf{Embedding}
&
&
&
&
\\
\hline

Image-only &
$\times$ &
$\times$ &
-- &
$\checkmark$ & 0.154
 & 0.335
 & 0.390
 & 0.285 
 \\

Cross-Attention Fusion &
$\checkmark$ &
$\checkmark$ &
Cross-Attn &
$\checkmark$ & 0.139
 &  0.332 
 &  0.382
 & 0.286
 \\

Without State Embedding &
$\checkmark$ &
$\checkmark$ &
Add + LayerNorm &
$\times$ & 0.108
 & 0.303
 & 0.361
 & 0.249
 \\

TransXplainRRG (Full) &
$\checkmark$ &
$\checkmark$ &
Add + LayerNorm &
$\checkmark$ & \textbf{0.228}
 & \textbf{0.438}
 & \textbf{0.655}
 & \textbf{0.337}
 \\

\hline
\end{tabular}
\end{adjustbox}
\end{table*}

\subsection{Report Generation Results} 

\subsubsection{Quantitative Results}

\CM{We evaluate the proposed TransXplainRRG model using multiple quantitative metrics commonly adopted in automatic radiology report generation. Tables \ref{tab: Word-overlap}, \ref{tab: F1Metrics}, and \ref{tab: F1Metrics_IU-X-ray} present the comparative results using both traditional natural language generation metrics (BLEU-4, METEOR, ROUGE-L, CIDEr) and clinically oriented evaluation metrics (BERTScore, F1-ChexBERT, F1-RadGraph, and GREEN).}

For several baseline models, performance values are reported from previously published studies. In cases where the original works did not report all evaluation metrics or where models were not publicly available, the reported values are taken from the comparative benchmark analysis presented in \cite{nicolson2024longitudinal}. To enable consistent comparison within our experimental setting, we also evaluated the publicly available MAIRA-2 model weights on the updated 398-sample of MIMIC-CXR test set and computed the required metrics. The results of our proposed model, TransXplainRRG, are obtained using the same evaluation protocol and dataset. These conventions apply to all result tables in this section, and the specific source of each result is indicated in the corresponding table footnotes.

The results presented in Table~\ref{tab: Word-overlap} highlight the performance of various automatic report generation models evaluated on the MIMIC-CXR test set using standard word-overlap metrics: BLEU-4 (B4), METEOR, ROUGE-L, and CIDEr. \CM{Because prior studies report results on different MIMIC-CXR test subsets with varying evaluation configurations, this table should be interpreted primarily as a contextual benchmark across the literature. A strict comparison under identical testing conditions is performed using the re-evaluated MAIRA-2 model and our proposed TransXplainRRG.}\cdm{Under the controlled evaluation setting using the updated 398-study test subset, TransXplainRRG demonstrates strong performance across the evaluated metrics.} It achieves the highest CIDEr score (0.279), which is particularly notable as CIDEr emphasises consensus across reference reports and better captures clinically meaningful content.\cdm{This score is higher than the corresponding TransXplainNet result (0.219) and the literature-reported results for MedXChat (0.175) and $\mathcal{M}^2 \mathrm{fact}_{\text{ENTNLI}}$ (0.187). However, the latter results were obtained under different evaluation settings and are therefore included primarily as contextual references.}


\cdm{TransXplainRRG achieves a BLEU-4 score of 0.168, which is higher than TransXplainNet (0.145) under the same 398-study setting. The results for CXRMate (0.074) and UniXGen-256 (0.101) are included as contextual references because they were obtained using different test subsets and evaluation configurations.} This improvement suggests enhanced phrase-level overlap and structural alignment with the reference texts.\cdm{The METEOR score of 0.179 and ROUGE-L score of 0.330 further characterise the lexical and sequence-level similarity of the generated reports under this evaluation setting.}

\CM{MAIRA-2 was selected as a primary comparison baseline because its original work reported state-of-the-art performance across multiple evaluation metrics and publicly released inference model weights, enabling reproducible evaluation under a consistent experimental setup.} While the top BLEU-4 and METEOR scores are achieved by MAIRA-2 (0.231 and 0.417 respectively) on a different 2461-study test set, its performance drops significantly (BLEU-4: 0.035, CIDEr: 0.039) when re-evaluated on our 398-study subset without prior-study input, suggesting that the performance reported in the original MAIRA-2 evaluation benefits from the availability of longitudinal prior-study information. \cdm{This comparison highlights the sensitivity of report-generation performance to the availability of prior-study information and demonstrates the performance of TransXplainRRG in a single-study setting.}

\begin{table*}[htbp]
\caption{Quantitative comparison of state-of-the-art (SOTA) methods for automatic chest X-ray report generation on the MIMIC-CXR test set. \cdm{Note that the number of test samples and evaluation configurations varies across studies, making direct comparisons difficult.} Word-overlap metrics are reported, where B4 denotes the BLEU-4 score. \cdm{The best available results for each metric are highlighted in bold, while others are visualised using a color gradient.} For models that are not publicly available, we report their original evaluation results on the corresponding test sets and available metrics.}
\label{tab: Word-overlap}
\centering
\begin{threeparttable}
\begin{tabular}{|c|c|c|c|c|c|c|c|}
\hline
 \textbf{Models} & \textbf{Test Samples} & \textbf{Multi-Views}  & \textbf{B4} & \textbf{METEOR} & \textbf{ROUGE-L} & \textbf{CIDER} \\ \hline
                        
MedXChat\tnote{1} & 3858 images   & No  & \colorrampred{0.111}{0.035}{0.231} & \colorrampgreen{0.135}{0.118}{0.417}   & \colorrampred{0.264}{0.162}{0.390}  & \colorrampgreen{0.175}{0.039}{0.279}  \\

Nguyen et al. 2021 \cite{nguyen2021automated}          & 1000 Studies  & Yes  & \colorrampred{0.224}{0.035}{0.231}  & \colorrampgreen{0.222}{0.118}{0.417}   & \textbf{\colorrampred{0.390}{0.162}{0.390}}  & --        \\ 

UniXGen-256\tnote{1} & 2733 Studis   & Yes  & \colorrampred{0.101}{0.035}{0.231} & \colorrampgreen{0.156}{0.118}{0.417}   & \colorrampred{0.294}{0.162}{0.390}  & \colorrampgreen{0.138}{0.039}{0.279}  \\


$\mathcal{M}^2 \mathrm{fact}_{\text{ENTNLI}}$\tnote{1}  & 1624 studies   & Yes  & \colorrampred{0.083}{0.035}{0.231} & --- & \colorrampred{0.269}{0.162}{0.390}   & \colorrampgreen{0.187}{0.039}{0.279}
 \\ 
CXRMate 2024 \cite{nicolson2024longitudinal} & 1624 studies   & Yes  & \colorrampred{0.074}{0.035}{0.231} & \colorrampgreen{0.158}{0.118}{0.417}  & \colorrampred{0.255}{0.162}{0.390} & \colorrampgreen{0.172}{0.039}{0.279} 
                        \\
                        



                        
MAIRA-2 2024\cite{bannur2024maira} & 2461 studies  & Yes  & \textbf{\colorrampred{0.231}{0.035}{0.231}} & \textbf{\colorrampgreen{0.417}{0.118}{0.417} }  & \colorrampred{0.384}{0.162}{0.390}  & -- \\

TransXplainNet 2023 \tnote{2} \cite{mondal2023transformers} & 398 studies  & Yes  & \colorrampred{0.145}{0.035}{0.231} & \colorrampgreen{0.161}{0.118}{0.417}   & \colorrampred{0.310}{0.162}{0.390}  & \colorrampgreen{0.219}{0.039}{0.279} 
                        \\                                            
MAIRA-2\tnote{3} ~\cite{bannur2024maira} & 398 studies  & Yes  & \colorrampred{0.035}{0.035}{0.231} & \colorrampgreen{0.118}{0.118}{0.417}   & \colorrampred{0.162}{0.162}{0.390}  & \colorrampgreen{0.039}{0.039}{0.279} \\

TransXplainRRG & 398 studies  & Yes  & \colorrampred{0.168}{0.035}{0.231} & \colorrampgreen{0.179}{0.118}{0.417}  & \colorrampred{0.330}{0.162}{0.390}  & \textbf{\colorrampgreen{0.279}{0.039}{0.279}}   
                        \\   
\hline
                        
\end{tabular}

\begin{tablenotes}
\item[1] Results quoted from \cite{nicolson2024longitudinal}.
\item[2] TransXplainNet evaluated on the updated 398-sample MIMIC-CXR test set.
\item[3] MAIRA-2 model weights evaluated on the same 398-sample test set without prior-study and comparison input.
\end{tablenotes}

\end{threeparttable}
  \vspace{-10pt}
\end{table*}



\cdm{Table~\ref{tab: F1Metrics} reports a comprehensive evaluation of recent state-of-the-art models across multiple metrics, including BERTScore-F1, F1-ChexBERT (Macro and Micro), F1-RadGraph, and the recently introduced GREEN metric. TransXplainRRG achieves strong performance across the evaluated metrics, obtaining the highest BERTScore-F1 among the reported results. While models such as MAIRA-2 and CXRMate demonstrate strong performance under specific metrics, TransXplainRRG provides competitive performance across both linguistic and clinically oriented evaluation measures.}

\cdm{We also evaluated the clinically-grounded GREEN metric to provide a complementary assessment of report quality under a controlled evaluation setting. Since GREEN was not reported for most existing methods, we generated reports using the publicly available MAIRA-2 model weights and computed GREEN with the same evaluation pipeline and test samples as TransXplainRRG. Under this controlled setting, TransXplainRRG achieved a higher GREEN score than MAIRA-2 (0.327 vs. 0.245), corresponding to an approximately $33.5\%$ relative improvement. This comparison is limited to MAIRA-2 and TransXplainRRG, for which GREEN was evaluated under the same setting.}

When evaluated on the updated 398-sample MIMIC-CXR test set, excluding prior studies, MAIRA-2’s performance declined substantially: a $45\%$ drop in F1-ChexBERT (Macro), $36.7\%$ in F1-ChexBERT (Micro), and $68.2\%$ in F1-RadGraph. \cdm{In contrast, TransXplainRRG was trained and tested without relying on prior study information and maintains consistent performance in this single-study evaluation setting.} Compared to models like CvT2DistilGPT2, $\mathcal{M}^2$ fact, CXRMate, and Med-PaLM, TransXplainRRG achieves higher or competitive scores across all evaluated dimensions, demonstrating its reliability in both linguistic quality and clinical accuracy.

\begin{table*}[htbp]

\caption{\cdm{Quantitative comparison of report generation performance using BERTScore, F1-ChexBERT, F1-RadGraph, and GREEN. GREEN was newly computed for MAIRA-2 and TransXplainRRG using the same test set and evaluation pipeline and is not used for direct comparison with the remaining baselines. The best available result is highlighted in bold, with higher colour intensity indicating higher scores.}}

\label{tab: F1Metrics}
\centering
\begin{threeparttable}
\begin{adjustbox}{width=\linewidth}
\begin{tabular}{|c|c|c|c|c|c|}
\hline
 \textbf{Models} & \textbf{BERTScore-F1} & \textbf{f1-ChexBERT(Macro)} & \textbf{f1-ChexBERT(Micro)}  & \textbf{F1-Radgraph} & \textbf{GREEN} \\ \hline
 
Med-PaLM M \cite{tu2024towards} & --  & \colorrampgreen{0.398}{0.229}{0.416} & \colorrampred{0.535}{0.368}{0.581} & \colorrampgreen{0.267}{0.110}{0.346}  & --    \\ 
                        
LLaVA-Rad \cite{zambrano2025clinically}     & --   & \colorrampgreen{0.395}{0.229}{0.416}    & \colorrampred{0.573}{0.368}{0.581} & -- & -- \\ 

CXRMate 2024 \cite{nicolson2024longitudinal} & \colorrampred{0.700}{0.553}{0.704}  & \colorrampgreen{0.357}{0.229}{0.416} & -- & \colorrampgreen{0.272}{0.110}{0.346} & --    \\

CvT2DistilGPT2\tnote{1} & \colorrampred{0.596}{0.553}{0.704}  & \colorrampgreen{0.258}{0.229}{0.416} & -- & \colorrampgreen{0.226}{0.110}{0.346} & --    \\

$\mathcal{M}^2 \mathrm{fact}_{\text{ENT}}$\tnote{1} &  \colorrampred{0.553}{0.553}{0.704} & \colorrampgreen{0.235}{0.229}{0.416} & -- & \colorrampgreen{0.237}{0.110}{0.346} & --    \\

$\mathcal{M}^2 \mathrm{fact}_{\text{ENTNLI}}$\tnote{1} &  \colorrampred{0.681}{0.553}{0.704} & \colorrampgreen{0.311}{0.229}{0.416} & -- & \colorrampgreen{0.320}{0.110}{0.346} & --    \\

MAIRA-2\cite{bannur2024maira} &  --  & \textbf{\colorrampgreen{0.416}{0.229}{0.416}} & \textbf{\colorrampred{0.581}{0.368}{0.581}} & \textbf{\colorrampgreen{0.346}{0.110}{0.346}}  & --    \\

MAIRA-2\tnote{2} & \colorrampred{0.652}{0.553}{0.704}  &  \colorrampgreen{0.229}{0.229}{0.416} & \colorrampred{0.368}{0.368}{0.581} & \colorrampgreen{0.110}{0.110}{0.346} & \colorrampred{0.245}{0.245}{0.327}   \\           
TransXplainRRG \tnote{3}  & \textbf{\colorrampred{0.704}{0.553}{0.704}}  & \colorrampgreen{0.393}{0.229}{0.416} &  \colorrampred{0.559}{0.368}{0.581}  & \colorrampgreen{0.236}{0.110}{0.346}  & \textbf{\colorrampred{0.327}{0.245}{0.327}}

\\

\hline

\end{tabular}
\end{adjustbox}

\begin{tablenotes}
\footnotesize
\item[1] Performance values for baseline models are quoted from \cite{nicolson2024longitudinal}.
\item[2] \cdm{MAIRA-2 reports generated on the updated 398-sample MIMIC-CXR test set; metrics computed using our evaluation protocol.}
\item[3] The proposed model TransXplainRRG is evaluated on the same test set.
\end{tablenotes}

\end{threeparttable}

  \vspace{-10pt}
\end{table*}


Table~\ref{tab: F1Metrics_IU-X-ray} presents a comparative analysis of our proposed model, TransXplainRRG, against recent state-of-the-art (SOTA) approaches on the IU X-ray dataset using both NLP-based and radiology-specific metrics. \cdm{TransXplainRRG achieves a GREEN score of 0.605 on the full IU X-ray test set. Since GREEN was not reported for the other baseline methods, we additionally evaluated the publicly released MAIRA-2 model under the corresponding evaluation setting, obtaining a GREEN score of 0.584. Thus, TransXplainRRG achieves a higher GREEN score than the re-evaluated MAIRA-2 model, although this comparison is limited to these two models. The GREEN metric provides a complementary assessment of the clinical consistency of the generated reports.}

For clinical NLG metrics such as F1-ChexBERT and F1-RadGraph, our model demonstrates competitive performance. While LLaVA-Rad achieves a higher micro F1-ChexBERT score (0.535 vs. 0.484), and $\mathcal{M}^2 \mathrm{fact}_{\text{ENTNLI}}$ leads in F1-RadGraph (0.334), our model maintains consistently strong results across the reported clinical metrics. MAIRA-2 shows the best performance in BLEU-4 and F1-ChexBERT (macro); however, when we evaluated its publicly released model weights on the IU X-ray dataset following its specified preprocessing requirements, the test set size was reduced from 3306 to 3199. Under this setting, its performance dropped significantly, with relative percentage reductions of $63.2\%$ in BLEU-4, $28.1\%$ in ROUGE-L, $41.1\%$ in F1-CheXBERT (macro), and $28.4\%$ in F1-RadGraph. \cdm{Notably, in this evaluation setting, TransXplainRRG outperforms the re-evaluated MAIRA-2 results across these metrics and achieves a GREEN score that is 0.021 higher (0.605 vs. 0.584), corresponding to a $3.6\%$ relative percentage improvement.}

In terms of general NLP metrics (BLEU-4 and ROUGE-L), our model delivers moderate performance on the full dataset. When trained and evaluated using the official IU X-ray train/test split, TransXplainRRG achieved BLEU-4 and ROUGE-L scores of 0.228 and 0.438, respectively. Importantly, the GREEN score remains relatively stable across the full dataset and official IU X-ray split (0.605 and 0.593, respectively), indicating similar clinical consistency across the two evaluation settings. While individual models demonstrate stronger performance on specific metrics, TransXplainRRG offers balanced performance across linguistic and clinically oriented evaluation measures.

\begin{table*}[htbp]
\caption{\cdm{Quantitative evaluation on IU X-ray using NLP-based and radiology-specific metrics. GREEN was newly computed for MAIRA-2 and TransXplainRRG under the corresponding evaluation settings and is not directly compared with other baselines. The best available result is shown in bold; darker gradients indicate higher scores.}}

\label{tab: F1Metrics_IU-X-ray}
\centering
\begin{threeparttable}
\begin{adjustbox}{width=\linewidth}
\begin{tabular}{|c|c|c|c|c|c|c|c|}
\hline
 \textbf{Models}  & \textbf{Samples} & \textbf{BLEU-4} & \textbf{ROUGE-L} & \textbf{f1-ChexBERT(Macro)} & \textbf{f1-ChexBERT(Micro)}  & \textbf{F1-Radgraph} & \textbf{GREEN} \\ \hline
                         
LLaVA-Rad \cite{zambrano2025clinically} & 2163  & --   & \colorrampgreen{0.253}{0.197}{0.316}    & -- & \textbf{\colorrampgreen{0.535}{0.375}{0.535}} & -- & -- \\ 

CXRMate 2024 \cite{nicolson2024longitudinal} & 3331 &  \colorrampred{0.046}{0.039}{0.117}  & \colorrampgreen{0.282}{0.197}{0.316} & \colorrampred{0.277}{0.188}{0.319} & -- & \colorrampred{0.291}{0.194}{0.334} & --    \\

CvT2DistilGPT2\tnote{1} & 3331 &  \colorrampred{0.039}{0.039}{0.117}  & \colorrampgreen{0.255}{0.197}{0.316} & \colorrampred{0.220}{0.188}{0.319} & -- & \colorrampred{0.258}{0.194}{0.334} & --    \\


$\mathcal{M}^2 \mathrm{fact}_{\text{ENTNLI}}$\tnote{1} & 3331 &  \colorrampred{0.066}{0.039}{0.117} & \textbf{\colorrampgreen{0.316}{0.197}{0.316}} & \colorrampred{0.234}{0.188}{0.319} & -- & \textbf{\colorrampred{0.334}{0.194}{0.334}} & --    \\

MAIRA-2\cite{bannur2024maira} & 3306 & \textbf{\colorrampred{ 0.117}{0.039}{0.117}}  & \colorrampgreen{0.274}{0.197}{0.316} & \textbf{\colorrampred{0.319}{0.188}{0.319}} & \colorrampgreen{0.525}{0.375}{0.535} & \colorrampred{0.271}{0.194}{0.334}  & --    \\

MAIRA-2\tnote{2} & 3199 & \colorrampred{ 0.043}{0.039}{0.117}  & \colorrampgreen{0.197}{0.197}{0.316} & \colorrampred{0.188}{0.188}{0.319} & \colorrampgreen{0.375}{0.375}{0.535} & \colorrampred{0.194}{0.194}{0.334}  & \colorrampgreen{0.584}{0.584}{0.605}    \\

TransXplainRRG\tnote{2} & 3337 & \colorrampred{0.064}{0.039}{0.117}  & \colorrampgreen{0.222}{0.197}{0.316} &  \colorrampred{0.234}{0.188}{0.319}  & \colorrampgreen{0.484}{0.375}{0.535}  & \colorrampred{0.234}{0.194}{0.334} & \textbf{\colorrampgreen{0.605}{0.584}{0.605}}
 \\ 
\hline
TransXplainRRG\tnote{3} & 668 & 0.228  & 0.438 &  0.182  & 0.655  & 0.337 & \colorrampgreen{0.593}{0.584}{0.605} 

\\ 


\hline

\end{tabular}
\end{adjustbox}

\begin{tablenotes}
\footnotesize
\item[1] Results quoted from \cite{nicolson2024longitudinal}.
\item[2] \cdm{Models trained on MIMIC-CXR and evaluated on the full IU X-ray dataset; MAIRA-2 uses 3199 samples after preprocessing.}
\item[3] Trained and evaluated on the official IU X-ray split. 
\end{tablenotes}
\end{threeparttable}

  \vspace{-10pt}
\end{table*}

\CM{To assess the statistical reliability of the observed performance differences, we conducted additional statistical analysis using the F1-ChexBERT (macro) metric. Model performance was evaluated across five independent runs for each of two learning rates while keeping other hyperparameters constant. A paired-sample \textit{t}-test was performed using IBM SPSS Statistics (version 30.0.0.0(172), 2024). The analysis yielded a statistically significant difference (\textit{P} = 0.015, $\textit{P} < 0.05$), with a 95\% confidence interval. These results indicate that the observed performance difference is unlikely to be due to random variation across runs.}

\subsubsection{Qualitative Results}
\label{Qualitative_report}  

\begin{figure}[b]
  \centering
\includegraphics[width=0.9\linewidth, height=0.45\textheight, keepaspectratio]{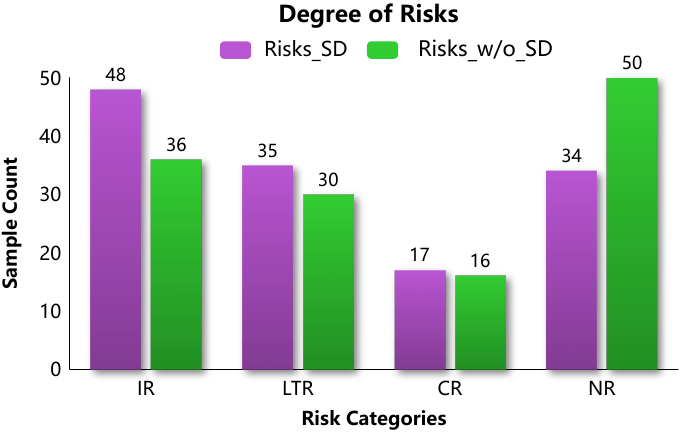}
  \caption{Qualitative assessment of generated reports based on primary evaluation by Radiologist-1. IR and LTR categories include reports classified as CR (both risks); categories are therefore not mutually exclusive. Inter-rater reliability with a second radiologist is reported separately.}
  \label{fig:Qualitative_evaluation}
\end{figure}

Figure \ref{fig:Qualitative_evaluation} illustrates the distribution of 100 chest X-ray reports generated by the TransXplainRRG model across four risk categories: Immediate Risk (IR), Long-term Risk (LTR), Combined Risk (CR), and No Risk (NR). These categories are analysed by considering support devices (Risks\_SD) and ignoring their role (Risks\_w/o\_SD). IR accounts for $48\%$ with support devices included, dropping to $36\%$ without, highlighting their significant contribution. LTR comprises $35\%$ with support devices and reduces slightly to $30\%$ without. CR, representing the overlap of immediate and long-term risks, is $17\%$ with devices and decreases to $16\%$ without. Lastly, NR cases make up $34\%$ with support devices and rise to $50\%$ without, aligning better with clinical safety when support devices are less emphasised. Among the 100 cases, only five were classified as normal findings reports, and the model correctly identified all of them, demonstrating its robustness in avoiding false negatives. By evaluating both Risks\_SD and Risks\_w/o\_SD distributions, this analysis highlights the model’s strengths in abnormality detection and areas for improvement, particularly in handling support devices and critical findings to enhance clinical safety.

\begin{figure*}[!ht]
  \centering
  \includegraphics[width=\linewidth]{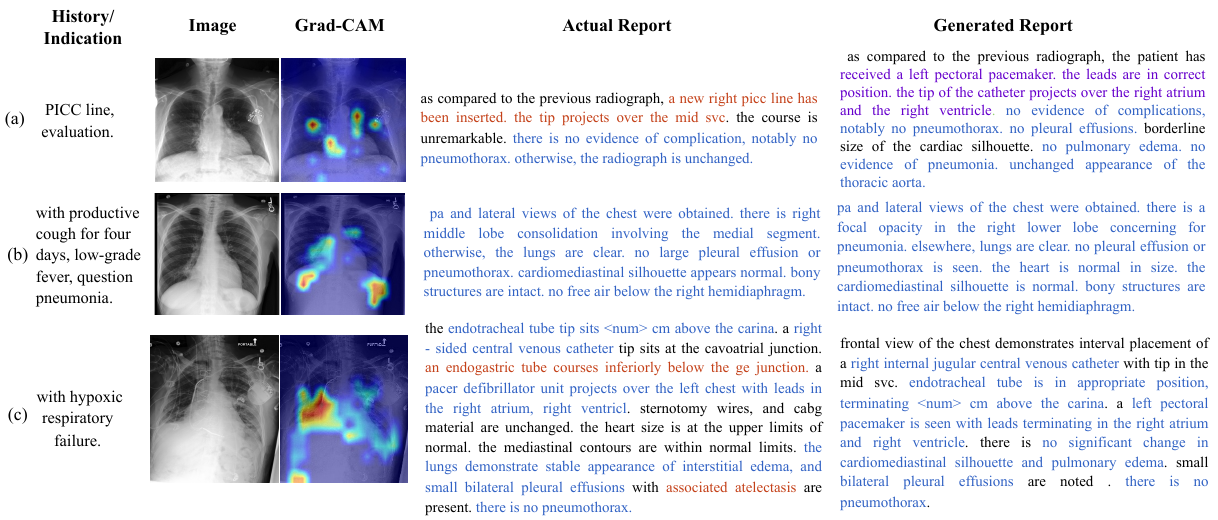}
  \caption{Qualitative representation of chest X-ray samples with history/indication, actual reports, generated reports, and heatmap visualisation. \textcolor{red}{Red} text indicates missed abnormalities, \textcolor{blue}{blue} text highlights correct findings, and \textcolor{violet}{violet} text represents correctly generated findings absent in the actual report.}
  \vspace{-10pt}
  \label{fig:Qualitative_report}
\end{figure*}

Figure \ref{fig:Qualitative_report} illustrates the qualitative analysis of reports of three sample test cases generated by TransXplainRRG. Each case is presented alongside the corresponding history, actual report, and generated report, with Grad-CAM heatmap visualisations used to assess the model’s performance. The analysis highlights both the strengths and limitations of the model in generating clinically meaningful reports. In the first sample, our model precisely detects abnormalities but misses reporting one support device (marked in red in the actual report). This omission is clinically considered an immediate risk. However, the model correctly identifies and describes a pacemaker visible in the X-ray but omitted in the actual report. Grad-CAM visualisation confirms the model's focus on the relevant region, highlighting its robustness in identifying missed findings. The second sample represents a high-quality report generated by the model, correctly addressing all findings with no risks identified. This demonstrates the model’s ability to produce clinically safe and comprehensive reports when the findings are well visualised and unambiguous on the chest X-ray. In the third sample, the model accurately reported three of four support devices and identified other abnormalities but missed one support device and atelectasis (highlighted in red). Missing a critical support device rendered this report both an IR and LTR, highlighting the need to improve the handling of complex and critical findings.

Overall, the qualitative assessment highlights the robustness of TransXplainRRG in generating clinically relevant reports across diverse cases. Its ability to detect abnormalities missed in the ground truth (violet text) and its accuracy in identifying multiple findings showcase its potential to support radiologists and improve workflow efficiency. However, the examples also emphasise the need for enhanced handling of critical support device reporting and rare findings to further strengthen its reliability and clinical applicability. 

\subsection{Analysis of LLM-Generated Customised Reports}

We generated reports using TransXplainRRG from chest X-ray images and clinical history and subsequently applied prompt-engineered LLMs to produce customised report representations. We investigated GPT-3.5-Turbo, GPT-4-Turbo, Llama-2-7B, Llama-2-13B, and Llama-3-8B using instruction-based, role-based, task-specific, contextual, and in-context prompting strategies. Two customised representations were considered: (i) an \textit{abnormal findings-only} report, which retains clinically salient abnormalities and support devices while removing routine normal findings, and (ii) an \textit{Inside-Out} report, which reorganises the generated findings according to a predefined anatomical and device-oriented sequence. The prompts were iteratively refined for task alignment and evaluated using MIMIC-CXR test samples.

\subsubsection{Study on Abnormal Findings}

We first investigated the ability of different LLMs to extract clinically relevant abnormal findings from the Stage I generated reports. An instructional information-extraction prompt was used, with the corresponding system prompts provided in Table~\ref{tab:Prompt_abnormal} of Appendix~\ref{Prompt_abnormal}. GPT-3.5-Turbo, GPT-4-Turbo, Llama-2-7B, Llama-2-13B, and Llama-3-8B were investigated using prompt engineering to obtain concise representations of clinically salient abnormalities. Based on the observed performance during this exploratory evaluation, GPT-4-Turbo and Llama-3-8B were selected for subsequent refinement experiments.

\subsubsection{Study on Inside-Out Approach}

We investigated an Inside-Out representation to impose a consistent anatomical organisation on the generated reports. The approach prioritises support devices and progressively describes the cardiomediastinal structures, lungs, pleura, and other relevant thoracic findings. GPT-4-Turbo and Llama-3-8B were selected for this analysis based on their observed performance in abnormal-finding extraction. Instruction-based prompting, contextual cues, chain-of-thought reasoning, and in-context learning were applied to GPT-4-Turbo, while prompt chaining was additionally incorporated for Llama-3-8B. The corresponding system prompts are provided in Table~\ref{tab:Prompt_Inside_Out} of Appendix~\ref{Prompt_abnormal}.

The Inside-Out approach was further investigated on 100 MIMIC-CXR test reports. The results of the prompt-engineering experiments were used to identify effective prompting strategies for report restructuring. In particular, prompt chaining was incorporated for Llama-3-8B to improve the consistency of the generated Inside-Out representations. GPT-4-Turbo and Llama-3-8B were therefore carried forward to the subsequent quantitative and qualitative evaluation of Stage II refinement.

\begin{cm}

\subsubsection{Quantitative Evaluation of Stage II Refinement}

To further assess the effect of LLM-based refinement, we conducted a paired before-and-after evaluation using the same 100 test reports. The original TransXplainRRG outputs were treated as Stage I reports, and the corresponding LLM-refined outputs were evaluated as Stage II reports. Both conventional text-generation metrics and clinical-content metrics were considered, including BLEU-4, ROUGE-L, CheXbert-F1, and F1-RadGraph. The results are presented in Table~\ref{tab:llm_refinement}.

\begin{table*}[t]
\centering
\caption{\cdm{Impact of LLM-based refinement on 100 test reports.}}
\label{tab:llm_refinement}
\begin{adjustbox}{width=\linewidth}
\begin{tabular}{|l|l|c|c|c|c|c|}
\hline
\textbf{Refinement} & \textbf{LLM/Stage} &
\textbf{BLEU-4} &
\textbf{ROUGE-L} &
\textbf{CheXbert Micro F1} &
\textbf{CheXbert Macro F1} &
\textbf{F1-RadGraph} \\
\hline

TransXplainRRG & Stage I &
0.159 & 0.300 & 0.563 & 0.399 & 0.323 \\
Inside-Out & Llama-3-8B &
0.071 & 0.273 & 0.568 & 0.402 & 0.303 \\
Inside-Out & GPT-4 Turbo &
0.053 & 0.257 & 0.559 & 0.397 & 0.285 \\

Abnormal Findings & Llama-3-8B &
0.007 & 0.132 & 0.523 & 0.369 & 0.136 \\
Abnormal Findings & GPT-4 Turbo &
0.003 & 0.142 & 0.537 & 0.379 & 0.134 \\

\hline
\end{tabular}
\end{adjustbox}
\end{table*}

The results show that Stage II refinement does not uniformly improve reference-based metrics. Compared with Stage I, both Inside-Out and Abnormal Findings-only refinement reduced BLEU-4 and ROUGE-L, reflecting the intended restructuring and compression of the reports. Inside-Out refinement with Llama-3-8B produced improvements in CheXbert Micro F1 (0.563 to 0.568) and Macro F1 (0.399 to 0.402), whereas GPT-4-Turbo yielded slightly lower scores. F1-RadGraph decreased for both LLMs, from 0.323 at Stage I to 0.303 and 0.285, respectively. Abnormal Findings-only refinement produced larger reductions in lexical and graph-based metrics, consistent with its selective removal of routine findings. Overall, Stage II is better viewed as a report-restructuring and information-prioritisation mechanism rather than as a component that universally improves reference-based metrics.

\subsubsection{Qualitative Analysis of Stage II Refinement}

Representative examples illustrating the benefits and limitations of Stage II refinement are provided in Table~\ref{tab:qualitative_llm_refinement} of Appendix~\ref{LLM-Refinement}. The Inside-Out strategy reorganises Stage I reports into a concise anatomical and device-oriented structure, while the Abnormal Findings-only strategy prioritises clinically salient abnormalities by removing routine negative observations. The examples also illustrate potential semantic and certainty-related errors introduced during refinement, including the conversion of stable pulmonary edema to an explicit absence and of uncertain findings to definitive abnormalities.

Overall, Stage II provides flexible, task-specific representations of Stage I reports. Inside-Out refinement promotes systematic anatomical organisation, whereas Abnormal Findings-only refinement prioritises clinically salient abnormalities. However, the observed metric changes and semantic errors indicate that automatic evaluation alone is insufficient to assess the clinical safety of LLM-based refinement, motivating further radiologist-based assessment of error severity and safety.

\end{cm}

\subsubsection{Overall Clinical and Qualitative Assessment}

To further examine the clinical reliability and safety of the refined reports, we analysed the radiologist-based assessment of the same 100 MIMIC-CXR test cases used in the Stage II evaluation. The assessment specifically examined the degree of error and the safety of the abnormal-findings-only and Inside-Out reports. The results are presented in Table~\ref{tab:Qualitative_Abnormal_Findings_inside_out}.

For the abnormal-findings-only representation, GPT-4-Turbo produced 88 error-free reports out of 100, followed closely by Llama-3-8B with 87. GPT-4-Turbo had 8 reports with minor errors (category 1), while Llama-3-8B had 10. Both models achieved a safety score of 100, indicating that no safety-related errors were identified in the evaluated outputs. In comparison, GPT-3.5-Turbo, Llama-2-13B, and Llama-2-7B produced fewer error-free reports and lower safety scores. These results indicate that GPT-4-Turbo and Llama-3-8B were able to produce concise abnormal findings-only representations with a favourable safety profile.

For the Inside-Out representation, the radiologist's assessment further highlights the influence of both model selection and prompt design on report restructuring. GPT-4-Turbo produced 92 error-free reports, with 5, 0, 2, and 1 reports assigned to error categories 1--4, respectively. Llama-3-8B initially produced 66 error-free reports, with 18 reports in category 1, 12 in category 2, 3 in category 3, and 1 in category 4. Notably, prompt chaining substantially improved the Llama-3-8B results, increasing the number of error-free reports from 66 to 85 and reducing the higher-severity errors. With prompt chaining, only one report was assigned to category 3 and none to category 4. These findings demonstrate that appropriate prompt engineering can substantially improve the reliability of LLM-based Inside-Out report restructuring, particularly for the smaller open-source Llama-3-8B model.

\begin{table}[htbp]
\vspace*{-5pt}
\caption{Qualitative assessment of Stage II report refinement applying prompt engineering based on evaluation by Radiologist-1. Inter-rater reliability with a second radiologist is reported separately. The evaluation metrics regarding degree of error and safety metrics are presented in 
Section \ref{EM_In_Out}.}
\label{tab:Qualitative_Abnormal_Findings_inside_out}
\centering
\begin{adjustbox}{width=\linewidth}
\begin{tabular}{|l||c|c|c|c|c|c|c||c|c|c|c|c|c|c|}
\hline
 & \multicolumn{7}{c||}{\textbf{Abnormal Findings}} & \multicolumn{5}{c|}{\textbf{Inside-out}}\\ \cline{2-13}
\cline{2-13}
\textbf{LLMs} & \multicolumn{5}{c|}{\textbf{Erroneous}} & \multicolumn{2}{c||}{\textbf{Safety}} & \multicolumn{5}{c|}{\textbf{Erroneous}}\\ \cline{2-13}
                  & \textbf{0} & \textbf{1} & \textbf{2a} & \textbf{2b} & \textbf{3} & \textbf{0} & \textbf{1} & \textbf{0} & \textbf{1} & \textbf{2} & \textbf{3} & \textbf{4}\\ \hline \hline
GPT-4-Turbo    & 88          & 8           & 1           & 3           & 0           & 0           & 100       & 92          & 5           & 0          & 2           & 1     \\ 
GPT-3.5-Turbo    & 78          & 17           & 0          & 4           & 1           & 2           & 98         & -& -& -& -&-  \\ 
Llama-2-13B    &     61       &  37         &   1         &     1       &    0       &  5         & 95       & -& -& -& -& -  \\ 
Llama-2-7B    & 53           & 31           & 0           & 12           & 4           & 4          & 96        & -& -& -& -&-   \\ 
Llama-3-8B    & 87           & 10           & 2           & 1           & 0           & 0           & 100       & 66           &   18         &  12          & 3           & 1     \\ 
Llama-3-8B  & -& -& -& -& -& -& -& 85           &   14         &  0          & 1           & 0                   \\
(chain-prompt)  & & & & & & & & &       &         &          &                   \\\hline
\end{tabular}
\end{adjustbox}
\vspace*{-10pt}
\end{table}

\subsection{Clinical Evaluation and Inter-Rater Reliability Analysis}

\CM{To reduce potential evaluation bias and assess the consistency of the clinical assessment criteria, a second radiologist evaluated the generated reports using the same risk and error definitions. Inter-rater reliability was measured using percentage agreement, Cohen’s $\kappa$, and quadratically weighted $\kappa$.} 

\CM{Table~\ref{tab:inter_rater_reliability} shows substantial agreement was observed for the inside-out report structuring task (weighted $\kappa$ = 0.713). For the generated report risk categorisation, moderate agreement was observed (weighted $\kappa$ = 0.446). The abnormal findings-only error-grading task achieved relatively high agreement (79\%), although the $\kappa$ value is reduced due to the strong class imbalance, with the majority of samples belonging to the no-error class. For the abnormal-findings-only safety assessment, both radiologists assigned identical labels to all cases, resulting in 100\% agreement and an undefined $\kappa$ value.}

\begin{table}[htbp]
\centering
\caption{Inter-rater agreement between two radiologists for clinical evaluation tasks. Agreement denotes identical labels; Cohen’s $\kappa$ and weighted $\kappa$ measure chance-adjusted agreement. The $\kappa$ value for abnormal findings-only safety evaluation is undefined because both radiologists assigned identical labels. 
}
\label{tab:inter_rater_reliability}
\setlength{\tabcolsep}{4pt}
\begin{adjustbox}{width=\linewidth}
\begin{tabular}{|l|c|c|c|}
\hline
\textbf{Evaluation Task} & \textbf{Agreement (\%)} & \textbf{Cohen’s $\kappa$} & \textbf{Weighted $\kappa$} \\ \hline

\makecell[l]{Generated report \\ risk assessment} & 55 & 0.303 & 0.446 \\ \hline

\makecell[l]{Inside-out report \\ structuring} & 92 & 0.468 & 0.713 \\ \hline

\makecell[l]{Abnormal findings-only \\ (degree of error)} & 79 & 0.095 & 0.054 \\ \hline

\makecell[l]{Abnormal findings-only \\ (safety)} & 100 & Undefined & -- \\ \hline

\end{tabular}
\end{adjustbox}
\vspace{-6pt}
\end{table}

\cdm{During the radiologist evaluation process, several common sources of error were observed, as summarised in Table~\ref{tab:error_analysis}. Support-device errors did not always arise from complete omission; in many cases, the model detected some devices but failed to report all devices when multiple devices were present in the same study. Errors in device presence, position, or laterality were also observed. When such omissions involved clinically important devices, the radiologists categorised them as potential risk factors.}

\cdm{Other observed errors included missed clinically relevant findings, incorrect anatomical localisation, and false-positive findings. For example, masses, nodules, and opacities were occasionally omitted, while some findings were assigned to an incorrect anatomical region or side. In a small number of cases, the model also introduced findings that were not supported by the reference report. These error patterns are summarised in Table~\ref{tab:error_analysis}. Inconsistent disease descriptions may also arise from the highly imbalanced distribution of the 14 disease categories in the MIMIC-CXR dataset, where certain findings appear far less frequently during training.}

\cdm{LLM-based refinement introduced additional variations in report wording and, in some cases, changes in the interpretation or certainty of findings. Representative examples of these effects, including both beneficial restructuring and errors introduced during refinement, are presented separately in Table~\ref{tab:qualitative_llm_refinement}. Overall, these observations highlight the importance of evaluating device reporting, abnormality detection, and anatomical localisation when assessing the reliability and safety of automatically generated reports.}

\begin{table}[t]
\centering
\caption{\cdm{Representative error categories observed in generated chest X-ray reports.}}
\label{tab:error_analysis}
\begin{tabular}{|p{0.21\linewidth}|p{0.7\linewidth}|}
\hline
\textbf{Error Category} & \textbf{Representative Observation} \\
\hline
Missed findings & Clinically relevant masses, nodules, opacities, fractures, or lung abnormalities were omitted. \\ \hline
Support-device errors & Device presence, position, or laterality was incorrectly reported or omitted, including NGT, PICC, ETT, and chest-tube information. \\ \hline 
Anatomical localisation errors & Findings were occasionally assigned to the wrong anatomical side or region. \\ \hline
False-positive findings & Findings not supported by the reference report were occasionally introduced, such as an absent port-a-cath or emphysema. \\ 
\hline
\end{tabular}
\end{table}

\section{Conclusion}

We developed TransXplainRRG, an enhanced chest X-ray report generation pipeline that advances research in this area. We identified Swin Transformer as an optimal vision encoder for disease classification and enabled explainability for radiologists by highlighting regions of interest during decision-making using Grad-CAM. Another key innovation was incorporating LLM-based refinement at the output stage to bypass fine-tuning costs while restructuring generated reports into clinically meaningful formats.
\cdm{On MIMIC-CXR, TransXplainRRG produced reports with clinically relevant structure and content}, supporting the inside-out reporting approach commonly used by radiologists and enabling abnormal findings-only summaries that can accelerate workflow. 
We found that small-scale open-source LLMs 
delivered an impressive performance with effective prompt engineering, offering privacy preservation and cost efficiency, making them \cdm{potentially suitable for privacy-sensitive hospital settings. Importantly, we proposed qualitative evaluation metrics from a radiologist’s perspective to assess the clinical utility of generated reports, providing expert-informed, risk-aware preliminary clinical assessment of report quality and potential clinical risks rather than clinical validation.} In future work, we plan to investigate bidirectional reasoning mechanisms, such as feedback loops between the LLM-based Stage II and the vision encoder in Stage I, to further refine attention localisation and reasoning. We also intend to evaluate the proposed approach on private hospital datasets with direct input from radiologists.
Additionally, we aim to further refine the LLM prompts to meet stringent privacy requirements, facilitating deployment on secure private networks. This direction will strengthen the practical applicability of our system and \cdm{support further investigation of its reliability across diverse clinical settings.}


\bibliographystyle{IEEEtran}
\small
\bibliography{ref}
\normalsize


\clearpage
\clearpage
\onecolumn

\appendix

\section{Prompt for Abnormal Findings Only and Inside-Out Report}
\label{Prompt_abnormal}
\noindent
\begin{table}[H]

\caption{Design of system prompts with GPT and LLaMA model to prepare abnormal findings only report.}
\label{tab:Prompt_abnormal}
\centering
\begin{adjustbox}{width=\textwidth}
\begin{tabular}{|p{0.07\linewidth}|p{0.93\linewidth}|} \hline
\textbf{Model} & \textbf{Prompt} \\
\hline
 GPT &  `` You are reading a radiology report. Extract concisely only the abnormal findings from the following report mentioned in user content. Any support devices like catheter, pacemaker, orthopedic hardware, wire should be considered as abnormal findings.\
Any recommendations or uncertainty related sentences or post operative status should be also considered as abnormal. Exclude any normal findings. If no abnormality is present, indicate Everything is Normal. 
" \\
\hline

LLaMA-2 & system prompts= [``Task: Extract Abnormal Findings and Support Devices related sentences.\
Scenario: You are analyzing a radiology report for a chest X-ray examination.\
Instructions:\
1. Extract only list of all the abnormal findings including support devices (e.g., catheters, sternotomy wires, ng tube, pacemakers) related sentences as concisely as possible.\
2. Exclude normal findings entirely.\
3. Include Recommendations, uncertainty-related sentences, or postoperative status related sentence.\
4. If the report states no abnormal findings, respond with only this single sentence: 'There are no abnormal findings'."]
\\
\hline

LLaMA-3 & ``role": ``system", ``content": ``You are analyzing a radiology report for a chest X-ray examination. Identify all abnormalities (including mild or subtle findings such as mild enlargement or changes in size, shape, or structure, post-operative status.) and support and monitoring devices (e.g., catheters, wires, pacemakers, NG tubes). \
Categorize your output into two sections only avoiding any note or extra writing :\
1. Abnormal Findings: Include all findings that suggest any deviation from normal, even if described as mild or subtle. If no abnormalities are present, just write 'No abnormal findings.\
2. Support and Monitoring Devices: Identify all devices mentioned in the report. If support devices are remain in place, mention that also."
\\
\hline

\end{tabular}
\end{adjustbox}
\end{table}



\begin{table}[H]
\caption{Prompts used with GPT and LLaMA-3 for Inside-Out report refinement. The chain-prompting variant applies a second prompt to further structure the reordered report.}
\label{tab:Prompt_Inside_Out}
\centering
\begin{adjustbox}{width=\textwidth}
\begin{tabular}{|p{0.1\linewidth}|p{0.85\linewidth}|} \hline
\textbf{Model} & \textbf{Prompt} \\
\hline
 GPT &  `` You are reading a radiology report. Rearrange the sentences of the findings within 1 paragraph maintaining the following order:
1. Any support devices like catheter, pacemaker, orthopedic hardware, wire-related sentences
2. Heart related sentences
3. Mediastinal or hili-related sentences
4. Lung related sentences
5. Chest wall or thoracic spine, or rib-related sentences
6. Abdomen related sentences
7. Rest sentences.
For your help,
any sentence related to the heart, cardio, pacemaker, atrium and ventricle is considered a heart-related sentence.
any sentence related to pleural effusion, pneumothorax, or pulmonary edema is considered a lung-related sentence.

Write exactly the same sentences. 
If any information about the order is missing, don't write anything about that.
" \\
\hline

LLaMA-3-8b & system prompts= [`` You are reading a radiology report as a normal reader. Your task is to rewrite the exact sentences of the report maintaining a sequential order and carefully following the instructions given below:

Instructions:
1. Write the exact sentences given in the report maintaining sequence of order. 
2. No extra sentences are allowed to be written.
3. Don't write like this: 'any of the orders are not mentioned'
4. No extra notes or introductory sentences are allowed.
5. If any of the orders are not mentioned in the report, just don't write anything about that.

Sequential orders:
1. Any support devices (includes monitoring and support devices, catheter, pacemaker, tube, wire, cabg) related sentences. 
2. Heart (includes heart, cardio, cardiac silhouette, cardiomegaly) related sentences. 
3. Mediastinal or hili-related sentences. 
4. Lung (includes pleural effusion, pneumothorax, or pulmonary edema, lung diseases, lung volumes) related sentences. 
5. Chest wall, thoracic spine or rib-related sentences. 
6. Abdomen-related sentences. 
7. Rest sentences.
]
"  \\
\hline

LLaMA-3-8b-chain-prompting & system prompts-1 =`` You are reading a radiology report. Your task is to reorder sentences into categories as instructed.

Instructions: Same as in the LLaMA-3-8b prompt.

Sequential orders: Same orders as the LLaMA-3-8b prompt. 
''


system prompts-2 = `` You are reading the reordered radiology report. Your task is to combine the exact sentences into a single paragraph without altering them.''
 \\
\hline

\end{tabular}
\end{adjustbox}
\end{table}

\clearpage

\section{LLM-Refinement Stage II Beneficials and Errors}
\label{LLM-Refinement}
\noindent

\begin{table}[H]
\centering
\caption{Representative examples of beneficial restructuring and potential errors in Stage II LLM-based report refinement.}
\label{tab:qualitative_llm_refinement}
\begin{adjustbox}{width=\linewidth}
\begin{tabular}{|p{0.08\linewidth}|p{0.06\linewidth}|p{0.3\linewidth}|p{0.26\linewidth}|p{0.28\linewidth}|}
\hline
\textbf{Strategy} &
\textbf{LLM} &
\textbf{Stage I} &
\textbf{Stage II} &
\textbf{Interpretation} \\
\hline

Inside-Out &
GPT-4 Turbo &
A right-sided port-a-cath tip sits in the lower SVC. The heart and mediastinal contours are normal. The lungs are largely clear with minimal right basal atelectasis and a small right pleural effusion. No pneumothorax. &
A right port-a-cath terminates in the mid SVC. Heart and cardiomediastinal contours are normal. The lungs are clear without focal consolidation. Small right pleural effusion. No pneumothorax. &
\textbf{Positive:} Concise anatomical/device-oriented reorganisation while retaining the principal findings. \\ \hline

Inside-Out &
GPT-4 Turbo &
No significant change in cardiomediastinal silhouette \textbf{and pulmonary edema}. Small bilateral pleural effusions. &
No significant change in cardiomediastinal silhouette. \textbf{No pulmonary edema.} Small bilateral pleural effusions. &
\textbf{Potential error:} Stable pulmonary edema is changed to an explicit absence of pulmonary edema. \\
\hline
Abnormal Findings-only &
Llama-3-8B &
Lungs are hyperinflated with flattened hemidiaphragms, consistent with COPD. Focal right lower lobe opacity concerning for pneumonia. Normal findings are also reported. &
\textbf{Hyperinflated lungs; flattened hemidiaphragms; focal right lower lobe opacity concerning for pneumonia.} &
\textbf{Positive:} Routine negative findings are removed while clinically salient abnormalities are retained concisely. \\
\hline
Abnormal Findings-only &
Llama-3-8B &
Residual patchy opacity \textbf{may reflect pleural effusion and collapse and/or aspiration} in the appropriate clinical setting. &
\textbf{Right upper lobe collapse; pleural effusion; aspiration.} &
\textbf{Potential error:} Uncertain findings are converted into definitive abnormalities. \\

\hline
\end{tabular}
\end{adjustbox}
\end{table}

\end{document}